\documentclass[conference]{IEEEtran}
\usepackage{stmaryrd}
\usepackage{amsfonts}

\usepackage{graphicx,times,epsfig,amsmath} 

\IEEEoverridecommandlockouts    

\begin{document}

\title{\ \\ \LARGE\bf A Data Mining Approach to Solve the Goal Scoring Problem\thanks{Renato Oliveira, Paulo Adeodato, Arthur Carvalho, Icamaan Viegas, Christian Di\^{e}go and Tsang Ing-Ren are with the Center of Informatics, Federal University of Pernambuco, Brazil (email: \{rmo, pjla, agc, ibvs, cdad, tir\}@cin.ufpe.br).} }

\author{Renato Oliveira \and Paulo Adeodato \and Arthur Carvalho
\and Icamaan Viegas \and Christian Di\^{e}go and Tsang Ing-Ren}


\maketitle

\begin{abstract}
In soccer, scoring goals is a fundamental objective which depends on many conditions and constraints. Considering the RoboCup soccer 2D-simulator, this paper presents a data mining-based decision system to identify the best time and direction to kick the ball towards the goal to maximize the overall chances of scoring during a simulated soccer match. Following the CRISP-DM methodology, data for modeling were extracted from matches of major international tournaments (10691 kicks), knowledge about soccer was embedded via transformation of variables and a Multilayer Perceptron was used to estimate the scoring chance. Experimental performance assessment to compare this approach against previous LDA-based approach was conducted from 100 matches. Several statistical metrics were used to analyze the performance of the system and the results showed an increase of 7.7\% in the number of kicks, producing an overall increase of 78\% in the number of goals scored.
 \end{abstract}


\section{Introduction}

\PARstart{T}{he} RoboCup organization was born with the objective of promoting artificial intelligence, robotics, machine learning and related fields using a benchmark to evaluate new technologies and algorithms. The soccer game was chosen as a central topic of research, due to its social characteristics and complex mechanics. The RoboCup project aims to promote innovative ideas that may be used to solve social and industrial problems.

Its main objective is to develop a robotic team composed of fully autonomous players that shall defeat the human world cup winners of the year 2050 \cite{cit:1}. Despite this ambitious objective, the Robocup project has been reaching some progress in several specific areas. In recent years, solutions for package routing in a network \cite{cit:2} and new algorithms for machine learning \cite{cit:3}\cite{cit:4} were proposed based on ideas inspired by the Robocup competition.

This competition was decomposed in several leagues, each league poses different challenges and offers varied levels of abstraction for the development of solutions. RoboCup Soccer 2D-simulator is one of these categories whose complex issues related to robotics, such as movement and computational vision, were simplified to a two dimensional space. In this category, each team is composed by twelve autonomous agents (eleven soccer players and one coach) acting on an environment with characteristics and constraints similar to those of the real game. The objective is to develop artificial intelligent solutions to the simulated game.

Since scoring goals is the main objective in a soccer game, one of the most important requirements to create a successful team in this environment is to have a clear policy whether the agent should attempt to score in a given scenario, and if so, which location in the goal the agent should aim for. The result of this action, i.e., goal scoring or not scoring, is influenced by a plethora of factors such as a good implementation of the opponent's goalkeeper, the specific position of ball, the randomness created by the simulator, the number of opponents near the ball and the goal, the shooting position of the player, the angle which the ball is kicked to the goal, among several other factors. Thus, the focus of this research is to propose a solution for this challenging problem. 

We propose a data mining solution developed according to the Cross Industry Standard Process for Data Mining (CRISP-DM) methodology \cite{cit:5}, where a decision system, based on neural networks, identifies the best moment and direction to kick the ball towards the opponent's goal. The modeling is based on data freely availability in the form of games played in previous competitions. 

The paper is organized in five more sections. Section 2 defines the problem of kicking towards the goal and its main characteristics. Section 3 describes the proposed solution following the steps suggested by CRISP-DM methodology. In this section it is showed how the data was collected, the preprocessing procedures and the transformation of variables used to embed the knowledge of the soccer specialist. Also we present a MultiLayer Perceptron (MLP) model developed to estimate the scoring chances and its performance on a statistically independent test set measured by ROC and KS2 curves. Section 4 describes how this decision support system was implemented combined with the team's strategy for kicking. Section 5 describes the experiments and compares the performance of the proposed approach to the best known solution in literature which is based on Linear Discriminant Analysis (LDA) \cite{cit:6}. Finally, Section 6 summarizes the results of the proposed approach and suggests additional topics for further research.

\section{Problem Characterization}
A soccer game presents complex situations where players need to take decisions about the actions that they have to perform. When they have the control of the ball, possible options could be, for instance, passing the ball to a teammate, dribbling an opponent or kicking the ball to the opponent's goal. In the last case, the agent has to decide before kicking whether the kick can give a positive return, i.e., increase the chance of scoring, based on his assessment on the current knowledge about the state of game. Particularly, the player carrying the ball has to decide, at each moment, if he will kick the ball or do another action according to the chances of scoring.

In the work \cite{cit:6} Kok \textit{et. al.} named this problem as {\em the optimal scoring problem} and they asserted that the problem could be decomposed into two independent subproblems:

\begin{enumerate}
    \item Determine the probability that the ball will enter the goal when shot to a specific point in the goal from a given position;
    \item Determine the probability that the ball will pass the goalkeeper in a given situation.
\end{enumerate}

It is interesting to note that the second subproblem can be seen as a two-class classification problem (goal or not goal). Since the subproblems are independents, the probability of scoring when shooting at a specific point in the goal will be equal to the product of these two probabilities.

Unfortunately, some assumptions were made in \cite{cit:6} to solve the second subproblem that weakened the final solution. First, the manner that the authors collected the data implied that the ball had to travel at least 20 meters before it reached the goal. Second, the possibility that other players besides the goalkeeper could be blocking the path to the goal was neglected. Finally, the data collected was composed by only two variables and a linear discriminant function, determined via linear regression (LDA) \cite{cit:7}, was used to separate the classes (goal or not goal).

Initially, to improve the solution, we reformulated the second subproblem as follows: {\em Determine whether kicking the ball in a specific position will result or not in a goal}. In this way, this subproblem turns now to be deterministic. Therefore, our final solution will combine the probabilistic solution to the first subproblem (proposed in \cite{cit:6}) with a deterministic solution, that is based on neural networks, which will solve this reformulated second subproblem.

Therefore, this work is focused on the problem of kicking the ball towards the opponent's goal in such a way that it does neither get intercepted by any opponent's defenders, including the goalkeeper, nor it goes outside the goal posts. This task is rather complex mainly for two reasons: the high complexity in predicting the actions performed by the opponents inside this Multi-Agent environment and the non-deterministic characteristics of the soccer simulator which mimics real factors.

The complexity related to the lack of knowledge about the opponent's behavior would lead, for instance, to kicking the ball towards the target whenever the chance of scoring was above a given threshold established based on a goalkeeper with average performance or a predefined rule. On one hand, if the goalkeeper performs much better than the average, the player will be kicking several times with much lower chances of scoring than estimated instead of taking another action that could increase the chances of success. On the other hand, if the goalkeeper performs much worse than the average, the player will be keeping the ball to increase the chances of scoring, instead of kicking it with chances higher than estimated.

The complexity related to the non-deterministic characteristics of the soccer simulator is due to the fact that the server disturbs the movement of the ball along the path traveled between the kick point and the target point on the opponent's goal. For each simulation cycle, a random value is added to the ball velocity vector, simulating the effects of a real environment (like wind). The model of the ball movement for each simulation step is described by the following equations \cite{cit:1}:

\begin{equation}
(u_x^{t+1}, u_y^{t+1})= (v_x^{t}, v_y^{t}) + (a_x^{t}, a_y^{t}) + (\widetilde{r}_{max}, \widetilde{r}_{max}) .
\end{equation}

\begin{equation}
(p_x^{t+1}, p_y^{t+1})= (p_x^{t}, p_y^{t}) + (u_x^{t+1}, u_y^{t+1}).
\end{equation}

\begin{equation}
(v_x^{t+1}, v_y^{t+1})= decay \times (u_x^{t+1}, u_y^{t+1}).
\end{equation}

\begin{equation}
(a_x^{t+1}, a_y^{t+1})= (0,0)
\end{equation}
where $(p_x^{t}, p_y^{t})$, $(v_x^{t}, v_y^{t})$ and $(a_x^{t}, a_y^{t})$ are vectors that represent respectively the position, the velocity and the acceleration of the ball in time step $t$ and $(u_x^{t+1}, u_y^{t+1})$ is the vector that represents the displacement in time step $t+1$. The decay factor is a deceleration constant and $r_{max}$ is a random value which makes the movement of the ball non-deterministic. The randomness factors are arbitrarily set by the simulator for each match and their values are not disclosed to the teams. At the end of each cycle the acceleration of ball is reseted. This value will be reassigned in the beginning of each cycle when the ball is kicked by some agent. The value for acceleration will change according to the power used in kicking.

Following the characterization of the optimal scoring problem, in the next section we will present a data mining approach to extract useful knowledge from scenes captured from official matches played in the major tournaments which are publicly available. This approach has the objective of improving the capacity of the players to make better decisions about kicking towards the goal, and thus solve efficiently the second subproblem. The combination of this new solution and the previous solution to the first subproblem will be explained in the section 4.

\section{The Data-Mining Based Solution}

According to the CRISP-DM methodology \cite{cit:5}, the process of data mining has six major phases: Business Understanding, Data Understanding, Data Preparation, Modeling, Evaluation, and Deployment.

In the following subsections we will explain how the data were collected and analyzed and how the model was constructed and evaluated. The phase of deployment will be explained in the section 4.

\subsection{Data Understanding}

The database with the goal scenes for mining the knowledge about goal scoring was obtained from matches played in the most important Robocup editions. These data are freely available and contain several results from teams of different levels. The data have been acquired at the exact moment that an agent kicks the ball towards the goal. A total of 10691 instances were extracted from tournaments played between 2005 and 2007 (we constrained data from these 3 years because of the evolution of teams in recent years) with a total number of 5853 goals scored and 4838 balls either caught by the goalkeeper or kicked outside the goal.

The 25 extracted variables have been chosen based on previous knowledge about the problem, taking their results in the real application domain (soccer) as metric. Some of these variables use information of the desired kick position in the opponent's goal. However, this information was not available at the moment of action and it has to be inferred from the angle used in the kick, the speed vector of the ball and the distance to the opponent's goal. The variables that have information about opponent's players considered only those who had simultaneously a X-axis position between attacker's X-axis position and the line of goal and had a Y-axis position within the goalkeeper's great area.

Some deficiencies in the match log recording process prevent this data mining approach from exploiting its entire potential. We do not know exactly when an agent kicks the ball towards the opponent's goal. To predict this fact, we look for the last kick given by a player before we got a message sent by the server informing that the goalkeeper caught the ball or the ball went outside the goal posts or that a goal was scored. This fact generates a problem when the ball is intercepted by a defender other than the goalkeeper. There is not a specific message sent by the server informing that this situation happened. In our work, these scenes have been discarded without loss of power of decision support once that proper care has been taken to include variables inside the model which measure, indirectly, the risk of interception.

\subsection{Data Analysis and Transformation of Variables}

After having the goal kicking scenes extracted from the soccer matches, an initial descriptive univariate statistical analysis was performed on each original input variable and new variables were derived to embed knowledge for systematically generating finer semantic information for decision making.

As each variable was numeric, the analysis consisted in measuring its expected value and standard deviation, median, first and last percentiles and the percentage of missing data. Univariate analysis on variables to approve the dataset was made. We treat missing, inconsistent and noise data removing outliers and adding specialist's knowledge systematically to generate more significant variables. At the end, we chose useful variables based on statistical metrics. Table 1 describes some of these variables.

\begin{table}[h]
\caption{Description of Some Variables}
\begin{center}
\begin{tabular}{|c|l|}
\hline
\multicolumn{1}{|c|}{Variable}
& \multicolumn{1}{|l|}{Description} \\
 \hline
                              & The angle formed between the line \\
                              & that goes from the attacker to the \\
Angle\_of\_Attacker's\_Vision & right side of the goal and the line \\
                              & that goes from the attacker to the  \\
                              & left side\\\hline
Angle\_of\_Players\_Vision    & Angle that determines the direction \\
                              & of attacker's vision \\ \hline
Goalkeepers\_X-Position       & X-axis position of the goalkeeper \\ \hline
Goalkeepers\_Y-Position       & Y-axis position of the goalkeeper \\ \hline
Ball\_X-Position              & X-axis position of the ball \\ \hline
Ball\_Y-Position              & Y-axis position of the ball  \\ \hline
                              & The angle formed between the line\\ 
Angle\_Ball\_Goalkeeper\_Destiny & that goes from the ball to the \\
                              & goalkeeper and the line that goes \\
                              & from the ball to the desired point\\ \hline
Strength\_of\_Kick            & The strength used by the player to \\
                              & kick the ball \\ \hline
Defender\_$i$\_X-Position     & X-axis position of the defender $i$ \\ \hline
Defender\_$i$\_Y-Position     & Y-axis position of the defender $i$ \\ \hline
Time                          & Time step of the kick \\ \hline
Results                       & The result of the kick \\ \hline
\end{tabular}
\label{tab-liu}
\end{center}
\end{table}

New variables have been created from initial vision of data to add useful knowledge in the system. Through the understanding of problem by specialist, it is possible to create many variables which have stronger semantics values than
the initial variables.

Some variables are irrelevant to the process of knowledge acquisition when their class distribution curves are similar. To detect these irrelevant variables we have generated The Receiver Operating Characteristic curves (ROC Curve) \cite{cit:8} for each variable. ROC Curve is a binary classification performance assessment tool which presents the relation between the true positive as a function of its false positive classifications produced for all the decision thresholds along the score continuous domain. The relevant aspect to be measured is the Area Under the Curve (AUC\_ROC) - a dissimilarity metrics whose maximum value is limited to one for a perfect classifier. Then, ROC curves supports the task of analyzing each variable relevancy. For this, each variable is assumed to be a classifier by itself, and the work is to analyze how much the ROC curve differs of a random classifier. In other words, we are looking for variables that have distant curves from the main diagonal of graph.

For instance, the variable {\em Angle\_Ball\_Goalkeeper\_Destiny} has an area on the curve of 0.793, so it was considered relevant. However, the variable {\em Time} obtained an area of approximately 0.504, therefore it was considered irrelevant. Although this analysis is univariate it is extremely useful to measure the relevance of variables related to the objective. A visual example of the curve for these variables can be seen in Figure 1.

\begin{figure}
\centering \epsfig{file=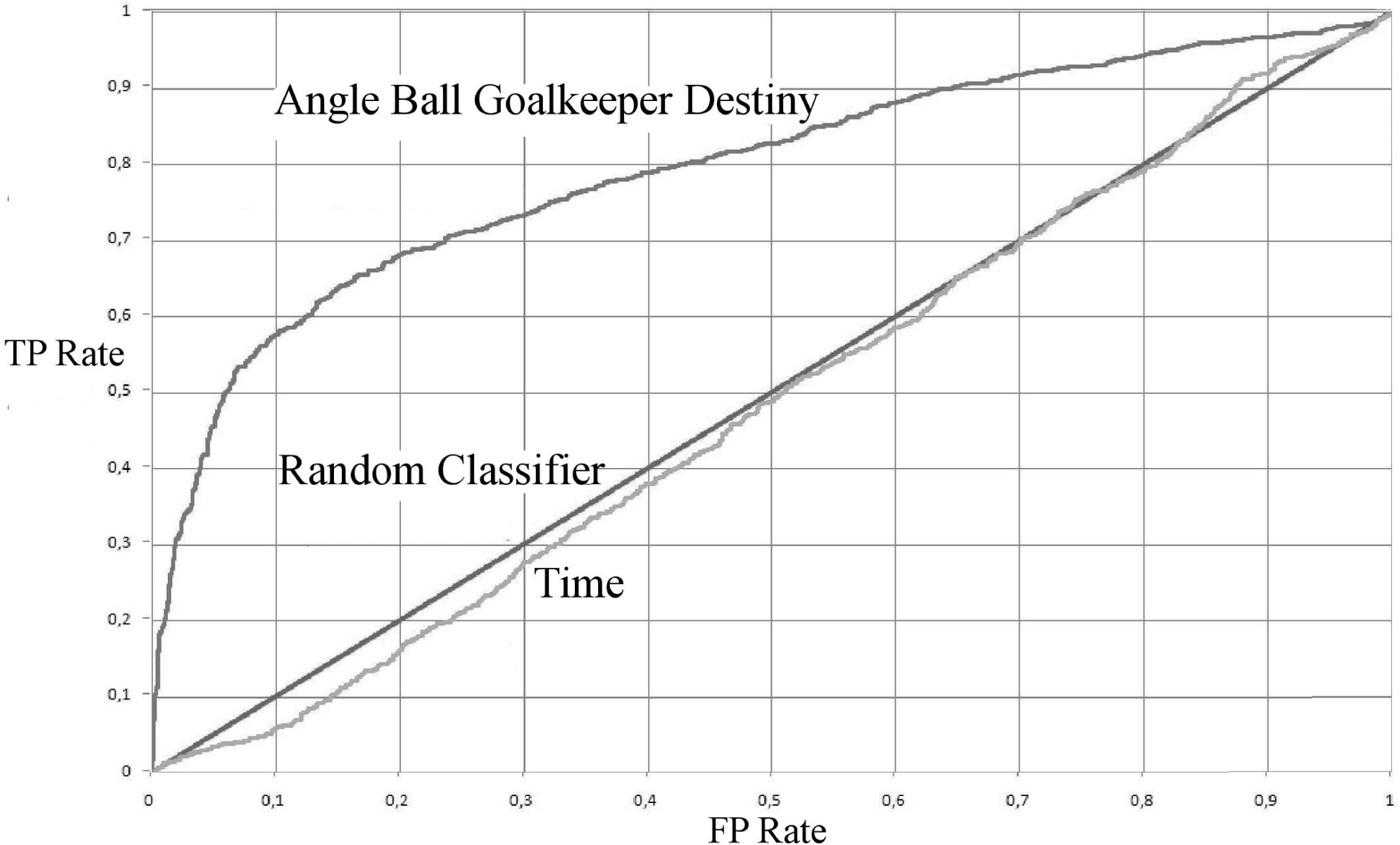,width=8.8cm} \caption{Roc Curve for Some Variables}
\end{figure}

According to the analysis of ROC curve, some variables present in the initial vision of data exhibited performances lower than their respective variables derived in the transformed vision. These variables and variables that had {\em posteriori} values, i.e. their values are defined after the kick, were discarded because they were considered irrelevant. At the end, 22 variables were used in modeling.

\subsection{Modeling and Evaluation}

The classifier selected to estimate the chances of scoring was the multilayer perceptron neural network (MLP) trained with the error backpropagation algorithm \cite{cit:9}. This selection was based on the several attractive features that this model possesses, for instance, its capacity of generalization, simplicity
of operation and universal approximation power.

Due to the necessity of a quick response in the real-time environment of the soccer simulator, we have chosen a simple network architecture with 22 nodes in the input layer (one for each variable), 5 nodes in the hidden layer (for smoothness of the result) and 2 nodes in the output layer (within the range $[-1,1]$) and used the hyperbolic tangent as the activation function.

The labeled data set was partitioned in training (50\%), validation (25\%) and test sets (25\%). Data present in training set were replicated for balancing the proportion of the two target classes. In this iterative training process the error for the backpropagation algorithm was measured by the mean squared error (MSE) \cite{cit:10} and the training stopping criterion was 10000 training epochs or five consecutive validation failures. The learning rate was constant and slow (0.001) and, to break symmetry, the connection weights before training were randomly set according to a uniform distribution ranging from $10^{-4}$ to $10^{+4}$. 

After learning, the MLP network was ready for performance assessment on the statistically independent test set (held out). First of all, we have transformed the output of the neural network using the following equation:

\begin{equation}
output = \left( \frac{node_1 - node_2}{4}\right) + 0.5 
\end{equation}

This continuous output within the range $ \left[0,1\right]$ allows a better performance control and assessment by setting a decision threshold applied to its scalar response for defining the kicking action based on the two possible results (classes) \cite{cit:11}: goal scoring or not scoring. Two metrics were used for the performance assessment on this continuous domain: the Kolmogorov-Smirnov (dis)similarity test (KS2) \cite{cit:12} and the Receiver Operating Characteristic Curve (ROC Curve) \cite{cit:8}. 

The KS2 \cite{cit:12} is a non-parametric statistical adherence test based on the difference of the cumulative distribution functions (CDFs) of the two classes in a binary decision problem. The relevant aspect to be measured is the maximum difference between the classes CDFs - a dissimilarity metrics whose maximum value is limited to one for a perfect classifier. Figure 2 shows the CDFs and the KS2 curve for the test set in which the KS2 metrics reached a maximum value of 0.740, indicating that the classifier produced a good level of separability between the distributions for the goal scoring and not scoring classes.

\begin{figure}
\centering \epsfig{file=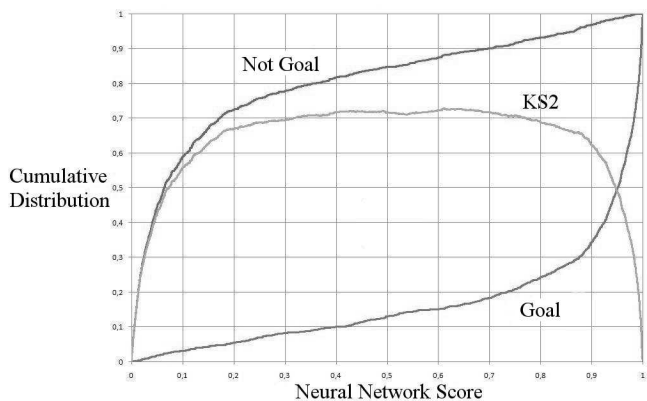,width=8.6cm} \caption{The KS2 curve for the test set}
\end{figure}

Figure 3 shows the ROC curve for the test set in which the AUC\_ROC metrics reached a maximum value of 0.941, also indicating that the classifier produced a good level of separability between the distributions for the goal scoring and not scoring classes.

\begin{figure}
\centering \epsfig{file=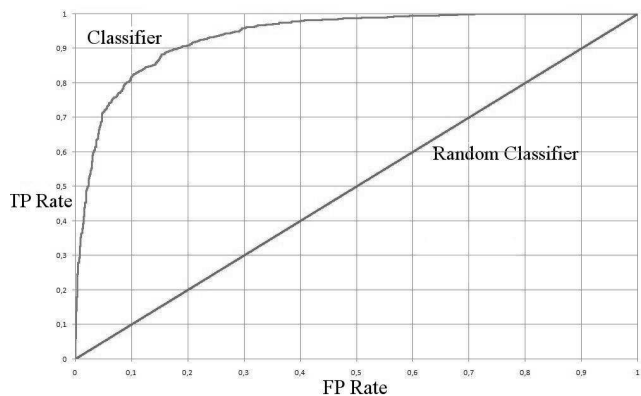,width=8.4cm} \caption{The ROC curve for the test set}
\end{figure}

\section{Implementation}

The proposed implementation consists of two main steps to solve the two subproblems. First of all, the opponent's goal has been discretized in fifteen  different target points uniformly distributed. To prevent extra processing and to solve the first subproblem, in the first step we use the method described by \cite{cit:6} to verify, for each target point in opponent's goal, the probability that the ball will not go outside the goal posts. This becomes necessary due to the fact that the noise added by server to the ball trajectory can impact on the result of the kick action.

Initially, Kok \textit{et. al.} have found that the standard deviation (with the center of the goal as mean) when the ball was kicked perpendicular to the desired point could be approximated by the function presented below:

\begin{equation}
f(d) = -1.88\times ln\left(1-\frac{d}{45}\right)
\end{equation}
where $d$ is the distance between the ball and the desired point in the opponent's goal. After that, the authors computed the distribution of the ball when it reaches the goal line and asserted that the cumulative noise will be approximately Gaussian. They found the likelihood that the ball goes out from left of the post, presented in the following equation:

\begin{equation}
P(\mbox{left}) \approx \int_{-\infty}^{S_l} \frac{1}{\sigma(d_l)\sqrt{2\pi}} \exp \left(-\frac{y^2}{2\sigma(d_l)^2} \right) dy
\end{equation}
where $d_l$ is the distance from the ball to the left post and $S_l$ is the shortest distance from the left goal post to the shooting line. Similarly, the likelihood that the ball goes out from right of the post is:

\begin{equation}
P(\mbox{right}) \approx \int_{S_r}^{\infty} \frac{1}{\sigma(d_r)\sqrt{2\pi}} \exp \left(-\frac{y^2}{2\sigma(d_r)^2} \right) dy
\end{equation}
where $d_r$ is the distance from the ball to the right post and $S_r$ is the shortest distance from the right goal post to the shooting line.

Based on above equations, the probability that the ball enters the goal will be $P(goal) = 1 - P(\mbox{left}) - P(\mbox{right})$. If this probability lies below a predefined threshold (empirically determined as 70\% estimated after some simulations) the option of kicking in that point will be discarded. Otherwise, a second step will be carried out on the resulting points.

In the second step, we get the score of the neural network (equation 5) for each valid point in the step one. If the result is greater than 0.5 (empirically defined after some simulations) then the agent will kick the ball in that point. Considering that more than one solution can be above this last threshold, the point with better result in the second step is chosen.

\section{Experiments}

The experiments were carried out on one hundred matches between two teams playing against each other, with implementation based on the work proposed by Boer and Kok \cite{cit:13}, but with only one difference between them: the algorithm for kicking towards the goal. The first algorithm was the solution proposed here and the other was the well-know known solution based on LDA proposed in \cite{cit:6}. The main metrics for performance assessment were the number of goals scored, the goal scoring success rate for each algorithm and the number of wins for each team. Table 2 shows the experimental results.

\begin{table}[h]
\caption{Comparison of Solutions on One Hundred Matches}
\begin{center}
\begin{tabular}{|c|c|c|}
\hline
\multicolumn{1}{|c|}{Metrics}
& \multicolumn{1}{|c|}{Neural Based Solution}
& \multicolumn{1}{|c|}{LDA} \\
 \hline
Kicks to goal & 279 & 259\\ \hline
Kicks (average per game) & 2.79 & 2.59\\ \hline
Kicks (standard deviation) & 1.701 & 1.577\\ \hline
Goals Scored & 96 & 54\\ \hline
Goals Scored (average per game) & 0.96 & 0.54\\ \hline
Goals Scored (standard deviation) & 0.875 & 0.673\\ \hline
Effectiveness & 0.344 & 0.208\\ \hline
Wins & 50 & 24\\ \hline
Losses & 24 & 50\\ \hline
Draws & 26 & 26\\ \hline
\end{tabular}
\label{tab-liu2}
\end{center}
\end{table}

The results presented show that the proposed algorithm is better than the LDA based approach in terms of goal scoring and, consequently, in terms of victories obtained. It is important to emphasize that in comparison to the LDA approach, the solution proposed here produces slightly more kicks towards the goal but its effectiveness is so much higher that it ends up with better results for goals scored.

\section{Conclusion}

This paper presented a data mining solution for the problem of when and where to kick the ball towards the opponent's goal ({\em optimal scoring problem}) in the environment of RoboCup Soccer 2D-simulator for maximizing the chances of scoring. Using data collected from many world top level competitions and following the CRISP-DM \cite{cit:5} process, this approach extracts significant knowledge from these data using a MLP neural network for assessing the chances of scoring goals. The experimental results have shown that the system performed extremely well when measured with KS2 and ROC Curve metric. When compared with the best known solution in literature \cite{cit:6}, we have obtained large advantage in the number of goals scored and by the goal scoring success rate.

Further improvements can be made if one can assess the chance of scoring by using the alternate option of not kicking towards the goal. This fact would help to define a threshold sensitive to the context to optimize the overall chance of scoring. This promising approach can also be used for other binary decision problems within the soccer playing application, particularly in optimizing the goalkeeper's actions because it could use the same scenes already captured for goal scoring, but this time with the opposite objective.

\IEEEtriggeratref{12}


%

\def\V{\rm vol.~}
\def\N{no.~}
\def\pp{pp.~}
\def\Pot{\it Proc. }
\def\IJCNN{\it International Joint Conference on Neural Networks\rm }
\def\ACC{\it American Control Conference\rm }
\def\SMC{\it IEEE Trans. Systems\rm , \it Man\rm , and \it Cybernetics\rm }

\def\handb{ \it Handbook of Intelligent Control: Neural\rm , \it
    Fuzzy\rm , \it and Adaptive Approaches \rm }

\end{document}